\documentclass[lettersize,journal]{IEEEtran}

\usepackage{cite}
\usepackage{amsmath,amssymb,amsfonts}
\usepackage{algorithmic}
\usepackage{graphicx}
\usepackage{textcomp}
\usepackage{booktabs} % For professional looking tables
\usepackage{comment}
\usepackage{amsmath}
\usepackage{algorithm}
\usepackage{algorithmic}
\usepackage{booktabs}
\usepackage{pifont} % for xmark
\usepackage{caption}
\usepackage{subcaption}
\usepackage{hyperref}
\usepackage{textcomp} %
\usepackage{fancyhdr}

\begin{document}

\pagestyle{fancy}
\fancyhead{}
\fancyhead[C]{Preprint submitted to Mechanical Systems and Signal Processing}

\title{EverAdapt: Continuous Adaptation for Dynamic Machine Fault Diagnosis Environments}

\author{Edward\textsuperscript{*}, Mohamed Ragab\textsuperscript{*}, \IEEEmembership{Member, IEEE},  Min Wu, \IEEEmembership{Senior Member, IEEE}, Yuecong Xu, Zhenghua Chen, Abdulla Alseiari, and Xiaoli Li,\IEEEmembership{Fellow, IEEE} 
\IEEEcompsocitemizethanks{
  \IEEEcompsocthanksitem{Edward was with the Centre for Frontier AI Research, Agency for Science, Technology and Research, Singapore (E-mail: edward003@e.ntu.edu.sg.}
    \IEEEcompsocthanksitem{Mohamed Ragab, and Min Wu are with the Institute for Infocomm Research, Agency for Science, Technology and Research, Singapore(E-mail: mohamedr002@ntu.edu.sg, \{wang\_yucheng, hou\_yubo, wumin\}@i2r.a-star.edu.sg).}
    \IEEEcompsocthanksitem{ Yuecong Xu is  with the Department of Electrical and Computer Engineering, National University of Singapore (NUS), Singapore (E-mail: yc.xu@nus.edu.sg.}
    \IEEEcompsocthanksitem{Abdulla Alseiari is with Propulsion and Space Research Center, Technology Innovation Institute, UAE
}
    \IEEEcompsocthanksitem{Zhenghua Chen and Xiaoli Li are with the Institute for Infocomm Research, Agency for Science, Technology and Research, Singapore, Centre for Frontier AI Research, Agency for Science, Technology and Research, Singapore (E-mail: chen0832@e.ntu.edu.sg, xlli@i2r.a-star.edu.sg).}  \IEEEcompsocthanksitem{\textsuperscript{*}These authors contributed equally to this work.}
}
}

\maketitle

\begin{abstract}

Unsupervised Domain Adaptation (UDA) has emerged as a key solution in data-driven fault diagnosis, addressing domain shift where models underperform in changing environments. However, under the realm of continually changing environments,  UDA tends to underperform on previously seen domains when adapting to new ones - a problem known as catastrophic forgetting. To address this limitation, we introduce the EverAdapt framework, specifically designed for continuous model adaptation in dynamic environments. Central to EverAdapt is a novel Continual Batch Normalization (CBN),  which leverages source domain statistics as a reference point to standardize feature representations across domains. EverAdapt not only retains statistical information from previous domains but also adapts effectively to new scenarios. Complementing CBN, we design a class-conditional domain alignment module for effective integration of target domains, and a Sample-efficient Replay strategy to reinforce memory retention. Experiments on real-world datasets demonstrate EverAdapt superiority in maintaining robust fault diagnosis in dynamic environments. Our code is available here: \href{https://anonymous.4open.science/r/EverAdapt-4876/README.md}{EverAdapt-Code}.

\end{abstract}

\section{Introduction}

\begin{figure}[ht!]
    \centering
    \includegraphics[width = \columnwidth]{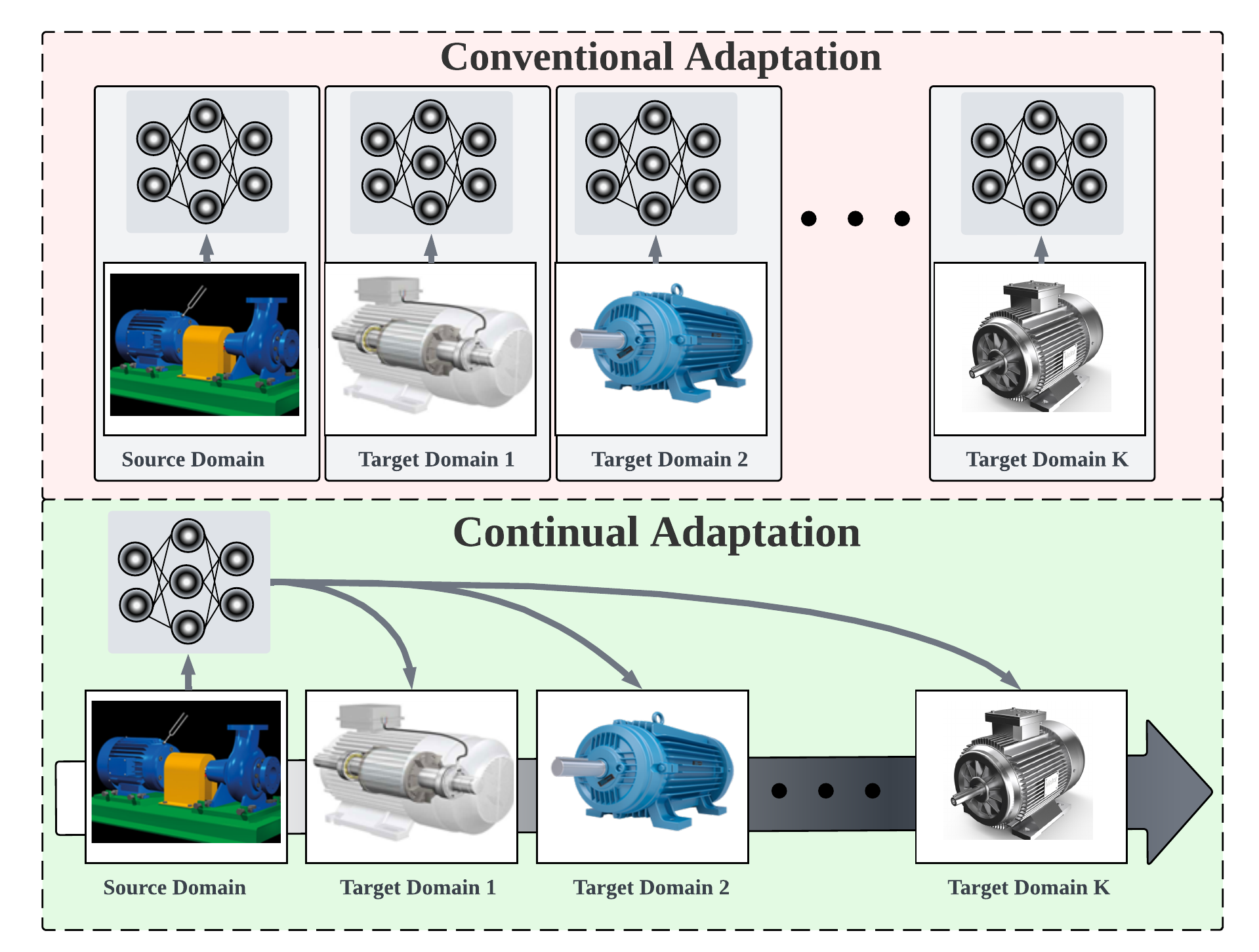}
    \caption{Comparison of Conventional and Continual Adaptation Approaches in Domain Adaptation. \textbf{Top}: the conventional adaptation approach, where individual models are independently trained for each new target domain. This often results in a scalability issue as the number of target domains increases, necessitating separate model and training phases for each domain. \textbf{Bottom}: the continual adaptation strategy, which employs a singular model that is sequentially adapted across multiple target domains. This method maintains knowledge from previous domains, effectively mitigating catastrophic forgetting and promoting model adaptation across a series of domain shifts.}
    \label{fig:continual_da_scenario}
\end{figure}

%% Background
In machine fault diagnosis, a critical challenge is the distribution shift problem, where the models' performances decline due to differences in training (source domain) and testing (target domain) data distributions \cite{diff_machines_da,cheng2020wasserstein}. Unsupervised domain adaptation (UDA) emerges as a promising solution for addressing distribution shift challenges in fault diagnosis. It leverages labeled data from a source domain, such as publicly available or simulated data, and unlabeled data from a target domain with a related but different distribution~\cite{fine_tune_fd,sae_da,ae_da1}. 

% UDA challenge under sequential scenarios
UDA's primary challenge in dynamic environments is its traditional focus on adapting to a single target domain. This limitation becomes especially apparent in scenarios where a model sequentially encounters multiple domains. In predictive maintenance, it is crucial for a fault diagnosis model, initially trained under specific pressure and temperature conditions of a particular machine, to be adaptable to varying working environments over time. While UDA enables the model to adjust to the most recent domain, this often results in the loss of proficiency in previously learned domains, a phenomenon known as catastrophic forgetting \cite{c_forget}. 
%% Naive solutions and the need for continual adaptation 
A naive solution to this problem would be to train a new model for each set of conditions, but this approach is impractical and resource-intensive for continuous operation, as illustrated in Figure \ref{fig:continual_da_scenario}. Therefore, there is a need for a model must continually adapt to new domains without losing its ability to perform in earlier ones~\cite{c_forget}.

%% Start by replay as the most common way to address forgetting
%% Limitation of replay as it requires a large sample size, is not scalable. 
%% We find that one key reason for such an issue is the BN, where the model is domain-specific and tends to overfit on the current domain, making it require a large number of samples of previous domains to retain its knowledge.
%% V1:
%%% Motivation for CBN:
Recently, continual unsupervised domain adaptation methods have gained traction by allowing models to adapt to new domains without forgetting previous ones \cite{conda,grcl,gfr}. However, the majority of existing methods are designed for computer vision applications, which may fail to perform well on time series data in machine fault diagnosis applications. Further, we argue that batch normalization (BN) can be detrimental to knowledge retention when adapting to new domains in fault diagnosis applications. Specifically, BN adjusts the model to the current domain's statistics, overlooking those from previous domains. This causes the model to specialize in the latest domain, impairing its performance on previously seen domains. 
% Proposed method story
To address this issue, the ``EverAdapt'' framework is designed for continual model adaptation across diverse domains while addressing the catastrophic forgetting problem. The framework features a class-conditional domain alignment (CCA) module for integrating new domains, aligning them with the source domain at the class-wise level. This ensures effective domain adaptation by addressing class misalignment, crucial for consistent performance across different conditions. To address the catastrophic forgetting problem, we develop a novel Continual Batch Normalization (CBN), which standardizes the batch statistics across different domains using fixed statistics from the source domain. This process ensures consistent feature representation, significantly reducing the risk of forgetting when adapting to new domains. However, resetting target domains to source statistics in CBN can lead to training instability due to domain distribution shifts. To counter this,  we reduce the uncertainty of the learned features by minimizing their conditional entropy. This approach helps mitigate the instability caused by the adaptation of batch statistics from various domains to the source statistics. 
Beyond adapting batch statistics across domains, our approach augments CBN with simple self-training using replay samples to align fine-grained classes between domains. Notably, integrating CBN significantly cuts down the number of replay samples required for effective self-training.

% In addition to adaptation batch statistics across domains, we complement CBN with a simple self-training of replay samples to align the fine-grained classes across domains. Interestingly, the integration of CBN allows us to significantly reduce the sample size needed for effective self-training.

% While CBN effectively adapts batch statistics across domains,  it can still be susceptible to forgetting, as we may need to align fine-grained classes from previous domains to the current domains. To address, this we introduce a simple replay loss with self-training. Unlike traditional replay techniques  that necessitate large number of replay samples,  the integration of CBN allows us to significantly reduce the sample size needed for effective self-training.

% While CBN effectively adapts batch statistics,  it can still be susceptible to forgetting due to ths shif class-conditional shifts encountered in previous domains. To counter this challenge, we propose self-training of the replay samples to complement our CBN approach. Traditional replay techniques often necessitate large number of replay samples, but the integration of CBN allows us to significantly reduce the sample size needed for effective self-training. This efficient combination effectively addresses class-conditional shifts, ensuring robust mitigation of forgetting with minimal replay requirements.

In summary, EverAdapt presents a scalable and efficient framework adept at navigating the dynamic complexities of machine fault diagnosis. The primary contributions of this approach are summarized as follows:
\begin{itemize}
    \item Forgetting Prevention Module: Introducing a novel CBN technique via standardizing batch statistics across domains using fixed statistics from the source domain. This approach preserves consistent feature representation and substantially mitigates the risk of forgetting.
    \item Flexible Everadapt Framework: Versatile adaptability of the Everadapt framework, accommodating a range of techniques for adaptation and replay, making it apt for various fault diagnosis scenarios.
    \item Empirical Validation: Demonstrated superiority of the proposed approach through experiments on real-world datasets, showcasing significant improvements over state-of-the-art methods and substantial mitigation of the forgetting issue.
\end{itemize}

\begin{figure*}
    \centering
    \includegraphics[width=0.9\textwidth]{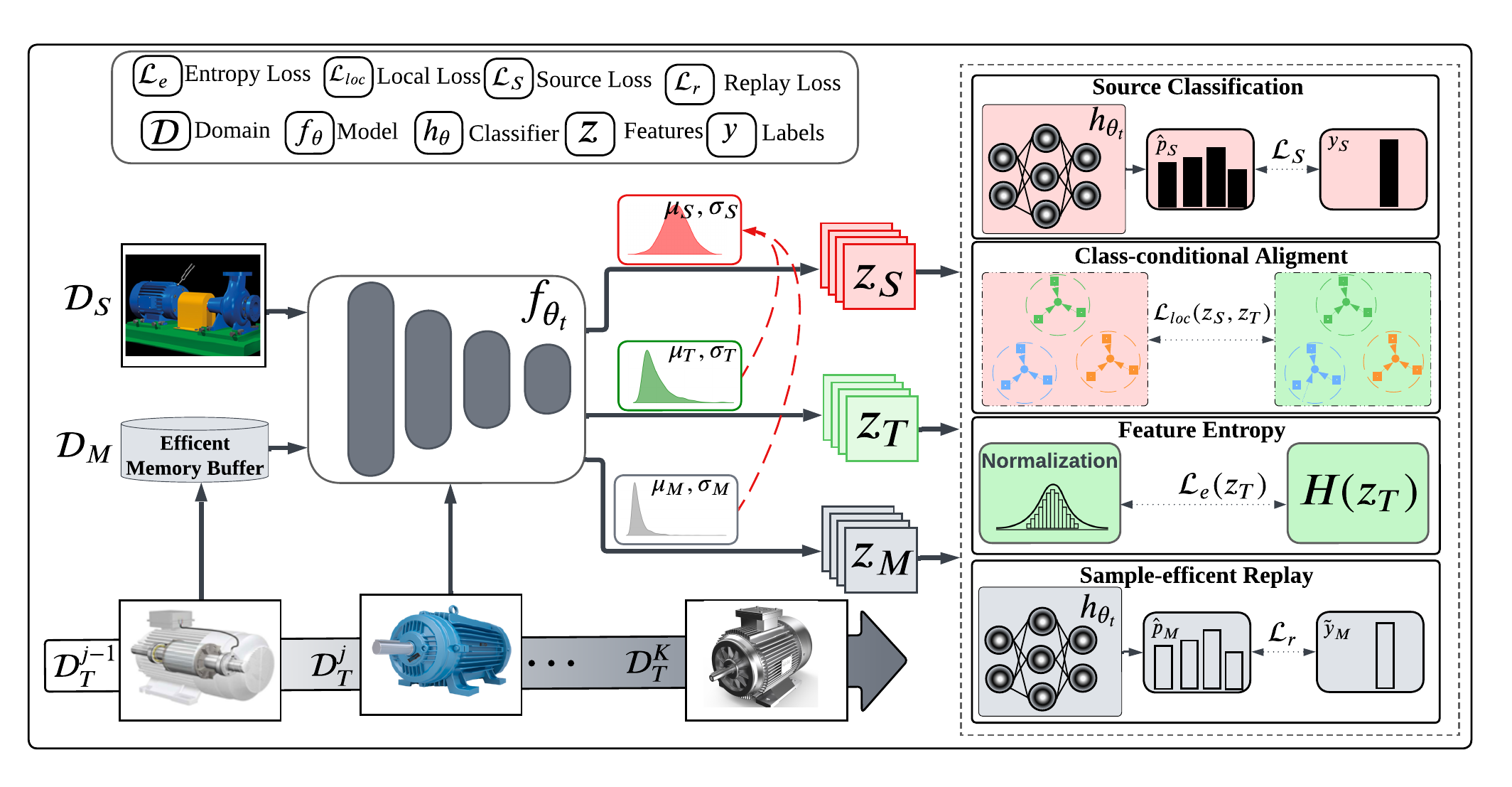}
    \caption{OverAdapt Framework Overview: incorporates input source samples, input target samples from the current target domain, and input memory samples to the feature extractor. It applies conditional entropy loss on the feature space of the target samples, cross-entropy loss on the input source samples, self-training with pseudo-labels on the memory samples, and local alignment loss between the source and target features.}
    \label{fig:EverAdapt}
\end{figure*}

\section{Related Works}
\subsection{Domain Adaptation for Fault Diagnosis}

In the field of machine fault diagnosis, domain adaptation has emerged as a vital solution for adapting models to diverse industrial environments. Early studies focused on aligning feature distributions using techniques like Maximum Mean Discrepancy (MMD) \cite{lu2016deep}. Adversarial networks were later introduced for improved distribution alignment \cite{li2019deep}. Recent advancements include class-conditional alignment methods \cite{wang2021subdomain}, which align not only feature distributions but also class-related information between domains. Some techniques leverage multiple source domains through weighting schemes \cite{zhu2020new}. While these approaches are effective in static environments with a single target domain, they encounter limitations when dealing with dynamic environments where models encounter multiple domains sequentially. Notably, as models adapt to new domains, they often suffer from the drawback of forgetting knowledge about previously encountered domains. This limitation underscores the need for novel methods to facilitate adaptation to sequential, dynamic domains while preserving knowledge from previous domains.

\subsection{Continual Domain Adaptation}

Continual adaptation to new domains while retaining knowledge of previous domains is a crucial challenge in computer vision applications. Existing methods have primarily focused on mitigating catastrophic forgetting when adapting to new domains. Feature replay has proven instrumental in addressing this problem, either through subsamples from previous domains \cite{cua,conda} or synthetic data generated by generative models \cite{grcl,frida}. Another approach involves parameter and weight regularization, achieved by either regularizing domain-specific features \cite{gsfda}, domain-specific neurons \cite{packnet}, or domain-specific weights \cite{eata}. While these methods have been effective in vision applications, they may not be directly applicable to signal data in machine fault diagnosis. Moreover, these approaches often overlook the contribution of Batch Normalization (BN) to the forgetting problem in previously seen domains. In contrast, we introduce a novel approach tailored to machine fault diagnosis. We present a simple yet effective Continual Batch Normalization (CBN) technique that addresses BN limitations and significantly reduces forgetting on previously seen domains

%\subsection{Partial Domain Adaptation} % Might not need to discuss this

\section{Methodology}
\subsection{Problem Definition}
% NEEDS CHANGE, need better wording & match the variables with equations
In the context of continual domain adaptation, we consider a source dataset \( D_S =\{{x}_S^i, {y}_s^i\}_{i=1}^{n_s} \) consisting of labeled samples, where each sample includes a signal \( {x}_S^i \) and a corresponding label \( y_{s}^i \). Moreover, we are presented with a sequence of target domains, denoted as \( \mathcal{D}_T = \{ D_T^1, D_T^2, \ldots, D_T^K \} \), each comprising unlabeled samples \( \{x_{T}^j\}_{j=1}^{n_T} \). The goal is to train a model \( f_\theta \) capable of accurately predicting labels across multiple target domains \( \{ D_T^1, \ldots, D_T^K \} \), each characterized by a unique marginal distribution \( P_{T}^i(x) \), distinct from the source domain's distribution \( P_S(x) \). The conditional distributions \( P(y|x) \) are assumed to be invariant across the source and target domains. The crux of the problem lies in training the model \( f_\theta \) not only to adapt to the distinct characteristics of each target domain but also to maintain and leverage the knowledge acquired from previous domains without the benefit of labeled data.

\subsection{Overview of EverAdapt}
EverAdapt integrates two key components: Class-Conditional Alignment (CCA), and Continual Batch Normalization (CBN) complemented with self-training, as illustrated in Figure \ref{fig:EverAdapt}. Specifcally, CCA effectively addresses domain shifts by maintaining fine structures during adaptation. Continual Batch Normalization, which normalizes incoming target domain data using source domain statistics, in conjunction with self-training of replay samples, ensures that the model retains knowledge of its previously learned domains without forgetting. The detailed algorithm is presented in Algorithm \ref{alg:continual_domain_adaptation}, and the subsequent sections provide a thorough discussion of each component.

%The EverAdapt approach integrates two complementary modules: the Dual-Adaptation Module and the Forgetting Prevention Module. The Dual-Adaptation Module efficiently handles domain shifts by aligning both global and local class features, ensuring effective adaptation to new domains. Complementing this, the Forgetting Prevention Module, featuring Continual Batch Normalization (CBN), normalizes new domain data using source domain statistics to stabilize feature representation. This is crucial for reducing catastrophic forgetting. Additionally, the inclusion of a sample-efficient replay mechanism within this module reinforces learning from past domains. Together, these modules enable EverAdapt to seamlessly adapt to new domains while retaining essential knowledge from previous ones. The detailed algorithm is illustrated in Algorithm \ref{alg:continual_domain_adaptation}
\subsection{Pretraining on Source Domain}
The source model architecture consists of a feature extractor \( f_{\theta_s}: \mathcal{X} \rightarrow \mathcal{Z} \), which maps the input space to the feature space \(\mathcal{Z} \in \mathbb{R}^d\), and a classifier \( h_{\theta_s}: \mathcal{Z} \rightarrow \mathcal{Y} \), responsible for mapping the feature space to class predictions. To train the source model, we utilize the standard cross-entropy loss \( \mathcal{L}_{ce}^s \), which is defined as:
\begin{equation}\label{eq:ce_loss}
\mathcal{L}_{ce}^s = -\sum_{c=1}^{C} \tilde{y}_{s,c} \log\left(p_{s,c}\right)
\end{equation}
where, \( p_{s,c} =  \sigma(h_{\theta_s}(f_{\theta_s}(x_S))) \) is the \( c \)-th element of the softmax output, \( \sigma(\cdot) \) represents the softmax function.

Once the source model is trained, we transfer its weights and batch normalization statistics to the target domains to obtain \( f_{\theta_t} \) and \( h_{\theta_t} \). This transfer sets the stage for training the target model to adapt sequentially to the incoming multiple target domains.

\subsection{Class-conditional Alignment (CCA)}
One of the key tasks in continual domain adaptation is the alignment of data distributions across different domains. However, conventional alignment methods primarily focus on aligning feature distributions between the source and target domains. While effective to some extent, they often overlook the fine-grained class distribution within each domain. This oversight can lead to a misalignment of similar classes across domains, negatively impacting the model's adaptation performance.
To address this challenge, we introduce our CCA module, which focuses on aligning class distributions between domains more granularly. Given the challenge of unlabeled target samples, our approach utilizes robust pseudo-labeling to classify target domain samples. Pseudo-labels are generated based on the highest probability class indicated by the model's predictions. The pseudo-label for a target sample \( x_T^j \) is given by:

\begin{equation}
    \hat{y}_T^j = \arg\max \sigma(f_{\theta_t}({z}_T^j)),
\end{equation}
where \( \hat{y}_T^i \) is the pseudo-label for the \( i \)-th target sample, and \( f_{\theta_t} \) represents the encoder model applied to current target domain time \( D_T^j \).
Once pseudo-labels are assigned, we align the class distributions by minimizing a class-level loss. This loss aims to reduce the discrepancy between the source and target distributions for each class. The class-level alignment loss \( \mathcal{L}_{loc} \)  can be expressed as:
\begin{equation} \label{eq:local_loss}
    \mathcal{L}_{loc} = \min_{\theta} \sum_{c=1}^{C} d(Z_S^c, Z_T^c),
\end{equation}
where \( C \) denotes the number of classes, \( Z_S^c \) and \( Z_T^c \) are the latent features for class \( c \) in the source and target domains, respectively. \( d(\cdot, \cdot) \) is a distance metric measuring the discrepancy between the two domains. Here, the Maximum Mean Discrepancy (MMD)  is employed as the distance between similar classes across domains, which defined as:
\begin{equation}
    \text{d}(Z_S^c, Z_T^c) = \left\| \mathbb{E}_{Z_S} [\zeta(Z_S^c)] - \mathbb{E}_{Z_T} [ \zeta(Z_T^c)] \right\|.
\end{equation}
In the above equation, \( \zeta \) is a feature map transforming the samples into a Reproducing Kernel Hilbert Space (RKHS) with a characteristic kernel \( k \), and \( \| \cdot \| \) denotes the norm in this space. The kernel function \( k \) is defined by the inner product in the RKHS: \( k(\cdot, \cdot) = \langle \zeta(\cdot), \zeta(\cdot) \rangle \). 

% Algorithm
\begin{algorithm}
\caption{Continual Domain Adaptation Algorithm}
\scriptsize
\label{alg:continual_domain_adaptation}
\begin{algorithmic}[1]
\REQUIRE Source dataset $\mathcal{D}_S$, sequence of target domains $\{\mathcal{D}_T^1, \mathcal{D}_T^2, \ldots, \mathcal{D}_T^K\}$.
\ENSURE Adapted model $f_\theta^K$ for the last target domain, performance metrics for the current domain, and backward transfer (forgetability) metrics for previous domains.
\STATE Pretrain the model $f_\theta$ on the source dataset $\mathcal{D}_S$.
\FOR{each target domain $t$ in $\{1, \ldots, K\}$}
    \STATE Input a batch of source samples from $\mathcal{D}_S$ into model $f_{\theta_t}$.
    \STATE Input buffer samples from the previous target domain $\mathcal{D}_T^{t-1}$ into model $f_{\theta_t}$.
    \STATE Input a batch of current target data from $\mathcal{D}_T^t$ into model $f_{\theta_t}$.
    \STATE Normalize the batch statistics of the current target domain and memory samples with respect to source statistics (refer to Eq.~\ref{eq:cbn}).
    \STATE Compute the source classification loss used during pre-training (refer to Eq. ~\ref{eq:ce_loss}).
    \STATE Compute the conditional entropy loss by minimizing the uncertainty of the target feature representation (refer to Eq.~\ref{eq:entropy_loss}).
    \STATE Compute the class-level alignment loss by minimizing the discrepancy between the source and target distributions for each class (refer to Eq.~\ref{eq:local_loss}).
    \STATE Optimize the models $f_{\theta_t}, h_{\theta_t}$ by minimizing the overall loss (refer to Eq.~\ref{eq:overall}).
    \STATE Assess performance on the current domain $\mathcal{D}_T^t$ post-adaptation.
    \IF{$t > 1$}
        \STATE Measure backward transfer (forgetability) on previous domains $\{\mathcal{D}_T^1, \ldots, \mathcal{D}_T^{t-1}\}$.
    \ENDIF
\ENDFOR
\STATE Evaluate the overall performance across all domains.
\end{algorithmic}
\end{algorithm}

\subsection{Preventing Catastrophic Forgetting}
% Introducing the challenge of forgetting
% Introducing conventional batch norm
% Introduce Continual Batch norm
A major challenge in continual adaptation is mitigating performance degradation on previously learned domains after adapting to new domains, a phenomenon known as catastrophic forgetting. In this work, we posit that batch normalization (BN) contributes significantly to this forgetting. To address this, we introduce a simple yet effective approach that adapts BN for sequentially arriving domains. We first discuss conventional BN to identify the underlying causes of forgetting. Subsequently, we present our CBN technique, designed specifically to overcome the issue of catastrophic forgetting in dynamic learning environments.

\subsubsection{Batch Normalization}
Batch Normalization (BN) is an essential technique in neural networks, aimed at addressing internal covariate shift. It normalizes the inputs of each layer to have zero mean and unit variance, contributing to the stabilization of the training process. For a mini-batch \(\mathcal{B}\), BN normalizes each input \(x_i\) as:
\begin{equation}
\hat{x}_i = \frac{x_i - \mu_{\mathcal{B}}}{\sqrt{\sigma_{\mathcal{B}}^2 + \epsilon}}.
\end{equation}
Here, \(\mu_{\mathcal{B}}\) and \(\sigma_{\mathcal{B}}^2\) are the mean and variance of the mini-batch, respectively, calculated by:
\begin{equation}
\mu_{\mathcal{B}} = \frac{1}{m} \sum_{i=1}^{m} x_i, \quad \sigma_{\mathcal{B}}^2 = \frac{1}{m} \sum_{i=1}^{m} (x_i - \mu_{\mathcal{B}})^2.
\end{equation}
The normalized input \(\hat{x}_i\) is then linearly transformed using learnable parameters \(\gamma\) and \(\beta\):
\begin{equation}
y_i = \gamma \hat{x}_i + \beta.
\end{equation}

A fundamental limitation of conventional BN in continual learning arises from its domain-specific normalization approach. BN normalizes inputs based on the current domain's statistics as:
\begin{equation}
    \text{BN}(x; \mu_{\text{domain}}, \sigma_{\text{domain}}^2) = \gamma \left( \frac{x - \mu_{\text{domain}}}{\sqrt{\sigma_{\text{domain}}^2 + \epsilon}} \right) + \beta
\end{equation}

In this context, \(\mu_{\text{domain}}\) and \(\sigma_{\text{domain}}^2\) are the mean and variance computed from the current domain's data. While this approach is effective for static data distributions, it can be problematic for continual learning. Rapid adaptation to the new domain's statistics (\(\mu_{\text{domain}}, \sigma_{\text{domain}}^2\)) may lead to a loss of information about previous domains' statistical properties, posing a challenge for models that need to perform well across diverse and evolving data streams.

\subsubsection{Continual Batch Normalization (CBN)}
To overcome the limitations of conventional BN in continual learning scenarios, we introduce CBN. This technique aims to preserve knowledge from previously learned domains while effectively adapting to new data, mitigating catastrophic forgetting. Unlike conventional BN, which recalculates mean and variance for each target domain, CBN standardizes the normalization process using statistics from the source domain.

During the source pretraining stage, we obtain running source statistics, including mean \(\mu_{\text{EMA}}\) and variance \(\sigma_{\text{EMA}}^2\), from each batch using Exponential Moving Average (EMA):
\begin{equation}
\begin{aligned}
\mu_{\text{EMA}} &= (1 - \alpha) \cdot \mu_{\text{EMA}} + \alpha \cdot \mu_{\mathcal{S}}, \\
\sigma_{\text{EMA}}^2 &= (1 - \alpha) \cdot \sigma_{\text{EMA}}^2 + \alpha \cdot \sigma_{\mathcal{S}}^2.
\end{aligned}
\end{equation}
Using these estimated source statistics, we standardize the batches of the all the incoming target domain:
\begin{equation} \label{eq:cbn}
\hat{x}_T = \frac{x_T - \mu_{\text{EMA}}}{\sqrt{\sigma_{\text{EMA}}^2 + \epsilon}}.
\end{equation}
By normalizing target domain data relative to the fixed statistics from the source domain, CBN maintains a consistent feature distribution across domains. This consistency ensures that knowledge from the source domain is preserved as the model adapts to new target domains, enhancing its generalization capabilities in continual domain adaptation tasks.

\subsubsection{Minimizing Features Entropy}
Resetting different target domains to the source statistics can cause instability in the training performance of CBN due to the distribution shift between domains. To address this, we aim to reduce the uncertainty of the learned features by minimizing their conditional entropy. This approach helps mitigate the instability caused by the differing adaptation of batch statistics from various domains to the source statistics. We formulate this process as follows: Given the target domain features \(z_T = f_{\theta_t}(x_T)\), we normalize these features to obtain \(\hat{z}_T = \text{Norm}(z_T)\). Finally, our objective is to minimize the conditional entropy of the normalized features, which can be expressed as:
\begin{equation} \label{eq:entropy_loss}
    L_e = \min_{\theta} H(\hat{z}_T | x_T),
\end{equation}
where \( H(\hat{z}_T | x_T) \) represents the conditional entropy, which quantify  the average uncertainty in the normalized feature set \( \hat{z}_T \) given the observed target data \( x_T \). By minimizing \( L_e \), we aim to reduce this uncertainty, thereby enhancing the features sharpness and, consequently, stabilizing the training process amidst varying domain-specific data distributions.

% Main benefits of CBN
%   - Better Adaptation [Need proof]
%   - Less Replay size requirement [See Replay Study figure]
% Therefore, introduce memory (intuitive way to reduce forgetting) -> CBN (reduce requirement)

\subsubsection{Sample-efficient replay}

While CBN can significantly reduce forgetting by referencing the batch statistics of incoming target domains to the statics of the source domain, there still exists a risk of forgetting due to variations in class distribution across different domains.  To address this, we enhance CBN with a simple replay method using a much smaller set of samples than conventional replay methods. This efficiency is mainly due to CBN's inherent capabilities. In our approach, replay samples are denoted as \(x_M\), with \(z_M\) representing their extracted features. The cross-entropy loss, \(\mathcal{L}_{ce}\) used for self-training the model with the predicted pseudo labels from  these replay samples:

\begin{equation} \label{eq:replay_loss}
    \mathcal{L}_{r} = \min_{\theta} \mathcal{L}_{ce}(h_\theta(f_\theta(x)_M), \tilde{y}_m)
\end{equation}
Here, \(\mathcal{L}_{r}\) represents the replay loss, \(h_\theta(z_M)\) are the predictions from the classification network for the features \(z_M\), and \(\tilde{y}_m\) are the corresponding pseudo labels for these replay samples.

\subsection{Overall Objective}
EverAdapt optimizes multiple objectives to facilitate adaptation to new domains while retaining knowledge from previous ones. These objectives include minimizing the conditional entropy of target features (\( \mathcal{L}_{e} \)), aligning source and target features with consideration for class information (\( \mathcal{L}_{\text{loc}} \)), self-training using memory samples (\( \mathcal{L}_{\text{m}} \)), and maintaining source classification performance (\( \mathcal{L}_{s} \)). However, balancing the minimization of entropy and class-conditional alignment (CCA) poses challenges, as excessive entropy reduction can result in prediction collapse into a single class, counteracting CCA's goal of precise class alignment across domains. To navigate this, we employ an adaptive weighting strategy. Initially, we prioritize entropy minimization (\( \mathcal{L}_{e} \)) with lesser emphasis on CCA loss (\( \mathcal{L}_{\text{loc}} \)). As training progresses, we gradually shift the focus, reducing entropy weight and enhancing the emphasis on CCA

% the concurrent optimization of CCA and conditional entropy can present challenges. Over-minimization of entropy, for instance, can lead to an undesirable collapse of predictions into a single class, undermining the core objective of CCA, which is to ensure a distinct and accurate alignment of classes across domains. To address this, we introduce an adaptive weighting technique. Initially, we assign a higher weight to conditional entropy (\( \mathcal{L}_{e} \)) at the start of training, with a smaller weight assigned to CCA loss (\( \mathcal{L}_{\text{loc}} \)). As training progresses and performance stabilizes, we dynamically adjust this balance, reducing the emphasis on entropy and increasing it on CCA.

The overall objective of EverAdapt is formalized as:

\begin{equation}\label{eq:overall}
\begin{split}
    \mathcal{L}_{\text{Overall}} = & \ \alpha(t) \mathcal{L}_{e} (z_T) + (1 - \alpha(t)) \mathcal{L}_{\text{loc}} (z_s, z_T) \\
    & + \beta \mathcal{L}_{\text{m}}(x_M) + \mathcal{L}_{s}(x_s,y_s)
\end{split}
\end{equation}

\begin{table}
  \centering
  \caption{PU dataset signal description}
  \label{tab:PU_description}
    \resizebox{\columnwidth}{!}{
  \begin{tabular}{lcccccc}
    \toprule
    Bearing & Damage level & Damage type & Location & Damage code & Type \\
    \midrule
    K001 & 0 & None & N/A & No Damage & Healthy \\
    \midrule
    KA01 & 1 & EDM & Outer & O-L1-EDM & Artificial \\
    KA03 & 2 & Engraving & Outer & O-L2-Engraving & Artificial \\
    KA05 & 1 & Engraving & Outer & O-L1-Engraving & Artificial \\
    KA07 & 1 & Drilling & Outer & O-L1-Drilling & Artificial \\
    KI01 & 1 & EDM & Outer & O-L1-EDM & Artificial \\
    KI03 & 1 & Engraving & Inner & I-L1-Engraving & Artificial \\
    KI07 & 2 & Engraving & Inner & I-L2-Engraving & Artificial \\
    \midrule
    KA04 & 1 & EDM & Outer & O-L1-EDM & Real \\
    KB23 & 2 & Engraving & Inner & I-L2-Engraving & Real \\
    KB27 & 1 & Engraving & Outer & O-L1-Engraving & Real \\
    KI04 & 1 & Drilling & Inner & I-L1-Drilling & Real \\
    \bottomrule
  \end{tabular}}
\end{table}

\begin{table}[hb]
  \centering
  \caption{Domain sequence used for each dataset}
  \label{tab:Scenario}
  \resizebox{\columnwidth}{!}{
  \begin{tabular}{cccccc}
    \toprule
    Dataset & Scenario & Source & Target 1 & Target 2 & Target 3 \\
    \midrule
    & 1 & A1 & A2 & A3 & A4  \\
    PU Artificial & 2 & A1 & A3 & A2 & A4 \\
    & 3 & A1 & A2 & A4 & A3  \\
    \midrule
    & 1 & R1 & R2 & R3 & R4  \\
    PU Real & 2 & R1 & R3 & R2 & R4 \\
    & 3 & R1 & R2 & R4 & R3  \\
    \midrule
    & 1 & U1 & U2 & U3 & U4 \\
    UO & 2 & U2 & U1 & U3 & U4 \\
    & 3 & U4 & U1 & U2 & U3   \\
    \bottomrule
  \end{tabular}
  }
  \label{tab:Domain_Order}
\end{table}

\begin{table}
  \centering
  \caption{Four working conditions of PU datasets, A/R denotes domains from PU Artificial and PU Real}
  \label{tab:PU_working_condition}
  \resizebox{0.5\textwidth}{!}{
  \begin{tabular}{ccccc}
    \toprule
    Domain & Rotating speed (rpm) & Load torque (Nm) & Radial force (N) \\
    \midrule
    A1/R1 & 1500 & 0.7 & 1000  \\
    A2/R2 & 900  & 0.7 & 1000\\
    A3/R3 & 1500 & 0.1 & 1000  \\
    A4/R4 & 1500 & 0.7 & 400   \\
    \bottomrule
  \end{tabular}}
\end{table}

\begin{table*}[htbp]
\centering
\caption{Comparative performance of our approach and baseline methods on the dataset across three distinct scenarios. Best results are denoted in bold while the second best are underlined.}
\resizebox{\textwidth}{!}{
\begin{tabular}{lccccccccc}
\toprule
\textbf{Methods} & \multicolumn{3}{c}{\textbf{PU Artificial}} & \multicolumn{3}{c}{\textbf{PU Real}} & \multicolumn{3}{c}{\textbf{UO}}\\
\cmidrule(r){2-4} \cmidrule(lr){5-7} \cmidrule(lr){8-10}\\
 & ACC & BWT & ADAPT & ACC & BWT & ADAPT & ACC & BWT & ADAPT \\
\midrule
CADA-DE \cite{yu2020conditional} & 80.94 $\pm$ 0.34 & -5.61 $\pm$ 0.42 & 84.67 $\pm$ 0.26 & 87.63 $\pm$ 1.35 & -10.15 $\pm$ 1.89 & 94.40 $\pm$ 1.12 & 78.37 $\pm$ 3.03 & -4.55 $\pm$ 3.31 & 81.40 $\pm$ 3.49\\
IDANN \cite{zhang2022multi} & 77.97 $\pm$ 0.92 & -7.01 $\pm$ 3.07 & 82.65 $\pm$ 2.24 & 91.62 $\pm$ 0.79 & -6.63 $\pm$ 1.52 & 96.04 $\pm$ 1.18 & \underline{84.68 $\pm$ 3.41} & -2.46 $\pm$ 4.96 & \textbf{86.32 $\pm$ 1.53} \\
HDDA \cite{wang2019hierarchical} & 82.18 $\pm$ 0.35 & -7.25 $\pm$ 1.52 & 87.01 $\pm$ 1.15 & 89.03 $\pm$ 1.56 & -10.75 $\pm$ 2.37 & 96.20 $\pm$ 1.21 & 80.38 $\pm$ 2.86 & -5.26 $\pm$ 5.10 & 83.89 $\pm$ 4.05\\
SATLN \cite{wang2021subdomain} &  81.44 $\pm$ 0.17 & -14.92 $\pm$ 0.93 & \underline{91.38 $\pm$ 0.64} & 94.14 $\pm$ 0.58 & -7.36 $\pm$ 0.80 & \underline{99.05 $\pm$ 0.16} & 81.98 $\pm$ 2.47 & -5.60 $\pm$ 6.60 & \underline{85.71 $\pm$ 4.46} \\
MMDA \cite{azamfar2020deep} & 82.04 $\pm$ 0.98 & -2.91 $\pm$ 2.57 & 83.98 $\pm$ 1.79 & 94.30 $\pm$ 0.56 & -3.72 $\pm$ 1.28 & 96.78 $\pm$ 0.77 & 79.41 $\pm$ 5.48 & -1.26 $\pm$ 5.49 & 80.26 $\pm$ 3.91\\ 
\midrule
ConDA \cite{taufique2021conda} & 75.71 $\pm$ 10.26 & -7.12 $\pm$ 5.72 & 80.46 $\pm$ 12.02 & 95.73 $\pm$ 1.20 & -5.54 $\pm$ 1.85 & \textbf{99.42 $\pm$ 0.17} & 62.98 $\pm$ 7.92 & -1.91 $\pm$ 6.88 & 64.26 $\pm$ 9.07 \\
CUA \cite{bobu2018adapting} & 83.46 $\pm$ 1.65 & -2.90 $\pm$ 1.42 & 85.39 $\pm$ 1.96 & \underline{96.69 $\pm$ 1.04} & \underline{-0.34 $\pm$ 0.80} & 96.92 $\pm$ 0.71 & 78.97 $\pm$ 9.85 & -1.95 $\pm$ 2.88 & 80.27 $\pm$ 8.83\\
DCTLN-DWA \cite{li2023deep} & \underline{84.14 $\pm$ 0.92} & \underline{-2.39 $\pm$ 0.44} & 85.73 $\pm$ 1.05 & 93.77 $\pm$ 0.85 & -1.40 $\pm$ 0.99 & 94.70 $\pm$ 0.80 & 83.18 $\pm$ 2.77 & \underline{0.01 $\pm$ 3.32} & 83.17 $\pm$ 3.28 \\
\midrule
EverAdapt & \textbf{92.81 $\pm$ 0.39} & \textbf{-1.10 $\pm$ 0.29} & \textbf{93.55 $\pm$ 0.88} & \textbf{99.05 $\pm$ 0.36} & \textbf{0.14 $\pm$ 0.31} & 98.96 $\pm$ 0.27 & \textbf{85.61 $\pm$ 4.49} & \textbf{0.34 $\pm$ 0.99} & 85.38 $\pm$ 4.19 \\
\bottomrule
\end{tabular}
}
\label{tab:PU baselines}
\end{table*}

\begin{table*}[htbp]
\centering
\caption{Ablation study of EverAdapt. 1\% was used as replay size}
\resizebox{\textwidth}{!}{
\begin{tabular}{ccccccccccccc}
\toprule
\multicolumn{3}{c}{\textbf{PU Artificial}} & \multicolumn{3}{c}{\textbf{Scenario 1}} & \multicolumn{3}{c}{\textbf{Scenario 2}} & \multicolumn{3}{c}{\textbf{Scenario 3}}\\
\cmidrule(l){4-6} \cmidrule(l){7-9} \cmidrule(l){10-12} 
CC & Replay & CBN  & ACC (\%) & BWT (\%) & ADAPT (\%) & ACC (\%) & BWT (\%) & ADAPT (\%) & ACC (\%) & BWT (\%) & ADAPT (\%) \\
\midrule
\checkmark & & & 81.14 $\pm$ 0.22 & -15.65 $\pm$ 2.79 & 91.76 $\pm$ 1.92 & 81.35 $\pm$ 0.23 & -18.84 $\pm$ 1.37 & \textbf{94.23 $\pm$ 0.95} & 82.08 $\pm$ 0.36 & -14.06 $\pm$ 2.63 & \underline{91.48 $\pm$ 1.84}\\
\checkmark & \checkmark & & \underline{85.27 $\pm$ 1.23} & \underline{-9.12 $\pm$ 1.86} & \underline{91.87 $\pm$ 2.08} & \underline{83.20 $\pm$ 0.32} & \underline{-9.19 $\pm$ 1.13} & 89.74 $\pm$ 0.82 & \underline{85.36 $\pm$ 0.98} & \underline{-8.37 $\pm$ 2.07} & 91.16 $\pm$ 2.00 \\
\checkmark & \checkmark & \checkmark & \textbf{93.11 $\pm$ 2.31} & \textbf{-1.68 $\pm$ 0.64} & \textbf{94.69 $\pm$ 2.08} & \textbf{91.56 $\pm$ 0.41} & \textbf{-1.04 $\pm$ 0.36} & \underline{92.88 $\pm$ 0.35} & \textbf{94.07 $\pm$ 0.98} & \textbf{-0.73 $\pm$ 0.51} & \textbf{94.67 $\pm$ 1.01}\\
\bottomrule
\end{tabular}}
\label{tab:Ablation study}
\end{table*}

\section{Experimental Settings}
\subsection{Dataset}
We validated our method using the Paderborn University (PU) bearing dataset and the University of Ottawa (UO) bearing dataset, which are ideal for testing a CDA setting due to its various working conditions. Details regarding each dataset will be discussed in the next section. Following the approach suggested by Zhao et al. \cite{zhao2020deep}, we used data segmentation to increase the size of both dataset and simplify the model's input requirements. Specifically, we applied a moving window technique with a window size and stride length of 1024 to segment the data, ensuring that the resulting data segments are distinct and non-overlapping for model training.

\subsubsection{Paderborn University Dataset}
The Paderborn University dataset \cite{lessmeier2016condition} contains vibration signals from an electric motor, with a total of 32 sets of signals, each representing a different bearing. Out of these, 6 bearings are healthy, 12 have artificial damage, and 14 have real damage from actual working conditions. Each bearing was tested under four different working conditions. Two dataset, named PU Artificial and PU Real, were created using combining signals from healthy bearings and artificially damaged bearings or bearings with real damage. Both subsets include a combination of healthy and faulty signals, as detailed in a table referred to as Table \ref{tab:PU_description}. In these datasets, the type of bearing is used as the class label and the different working conditions under which the bearings were tested are considered as different domains.

\subsubsection{University of Ottowa Dataset}
The University of Ottowa (UO) dataset \cite{uo_dataset} comprises vibration signals from bearings operating under varying health conditions and rotational speeds. A total of 36 set of signals are included, each corresponding to one of 12 experimental conditions derived from combinations of three bearing health states (healthy, inner race defect, outer race defect) and four rotational speed patterns (increasing speed, decreasing speed, increasing then decreasing speed, and decreasing then increasing speed). For each condition, three trials were conducted to ensure data reliability. In the UO dataset, the state of the bearing's health is used as a class label, and the different rotational speed patterns are considered as separate domains.

\subsection{Domain Scenarios}
We present the results of our method based on the average from three different scenarios for each dataset, as detailed in Table \ref{tab:Domain_Order}. This approach enhances the reliability of our results by preventing any bias towards specific scenarios that might favor certain methods.

% We tested our proposed EverAdapt method across three different scenarios to confirm its reliability in any domain order. These scenarios are detailed in Table \ref{tab:Domain_Order}. The results are an average of the outcomes from all three scenarios.

\subsection{Evaluation metrics}
We introduce three key metrics to assess a model's performance when adapting to multiple target domains. The first metric, average Accuracy (ACC), evaluates the model's overall performance across all observed domains. The second metric, average Backward Transfer (BWT), measures how well the model maintains its performance on previously adapted domains. The third metric, average Adaptation (ADAPT), assesses the model's effectiveness in adapting to unseen domains. Formally, we define \( R_{i,j} \) as the test accuracy on domain \( D_j \) after the model has adapted to domain \( D_i \). Here, \( N \) represents the number of target domains, and \( T \) denotes the total number of adaptation tasks. We can then express the calculations for the three metrics in the following equations:
% To discuss these metrics in detail, we define \( R_{i,j} \) as the test accuracy on domain \( D_j \) after the model has adapted to domain \( D_i \). Here, \( N \) represents the number of target domains, and \( T \) denotes the total number of adaptation tasks. The calculations for these metrics are summarized in the following equations
\begin{equation}
\text{ACC} = \frac{1}{N} \sum_{i=1}^{N} R_{N,i}
\end{equation}
\begin{equation}
\text{BWT} = \frac{1}{N - 1} \sum_{i=1}^{N-1} (R_{N,i} - R_{i,i})
\end{equation}
\begin{equation}
\text{ADAPT} = \frac{1}{N-1} \sum_{i=1}^{N} R_{i,i}
\end{equation}

\subsection{Implementation Details}
To ensure a fair comparison, all models, including EverAdapt, were assessed using a standardized feature encoder and classifier. The feature encoder comprises three convolution blocks, following the structure suggested by \cite{ragab2023adatime}. Key components of each block include a 1D convolution layer, batch normalization, a ReLU layer, and a max pooling layer. The first block features a 128-channel CNN layer with a kernel size of 5 and a dropout layer (dropout probability: 0.5). The second block doubles the channels, using a kernel size of 8, while the third block returns to 128 channels, also with a kernel size of 8. An adaptive layer then condenses the outputs to a length, leading into a fully connected classification layer.

Parameter settings were uniform across methods: a learning rate of $1 \times 10^{-3}$, weight decay of $1 \times 10^{-4}$, 40 epochs, and a batch size of 256. To validate robustness, each model underwent five runs with different random seed values, ensuring the reliability of the performance to seed variation.

\section{Results and Discussions}
\begin{figure*}
     \centering
     \begin{subfigure}[t]{0.32\textwidth}
    \includegraphics[width = \textwidth]{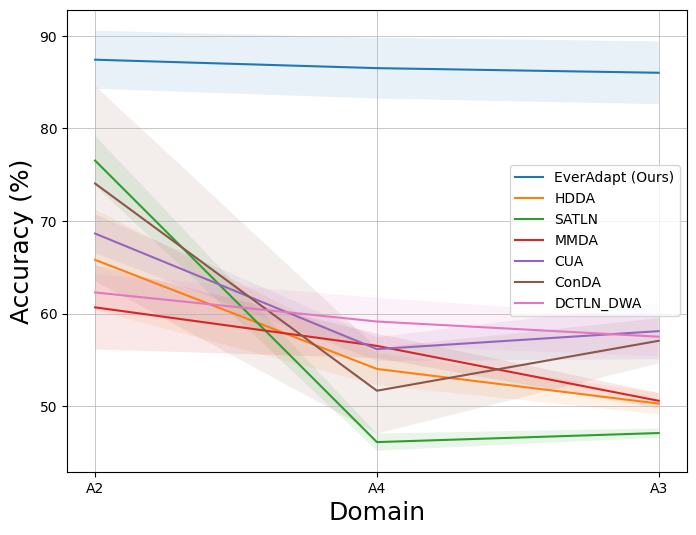}
    \caption{Forgetting performance on the first target domain A2}
    \label{fig:First_Target_Forgetting}
     \end{subfigure}
     \hfill
         \begin{subfigure}[t]{0.326\textwidth}
        \includegraphics[width = \textwidth]{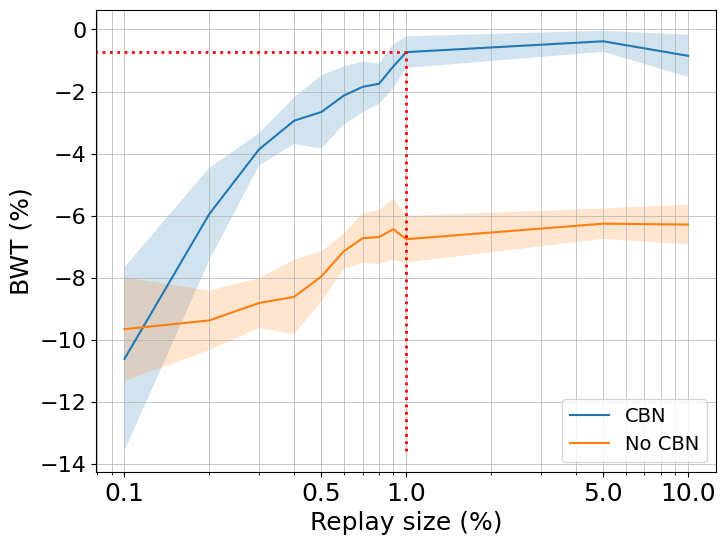}
        \caption{Performance of EverAdapt with different replay sizes}
        \label{fig:Replay_study}
     \end{subfigure}
      \hfill
    \begin{subfigure}[t]{0.327\textwidth}
    \includegraphics[width = \textwidth]{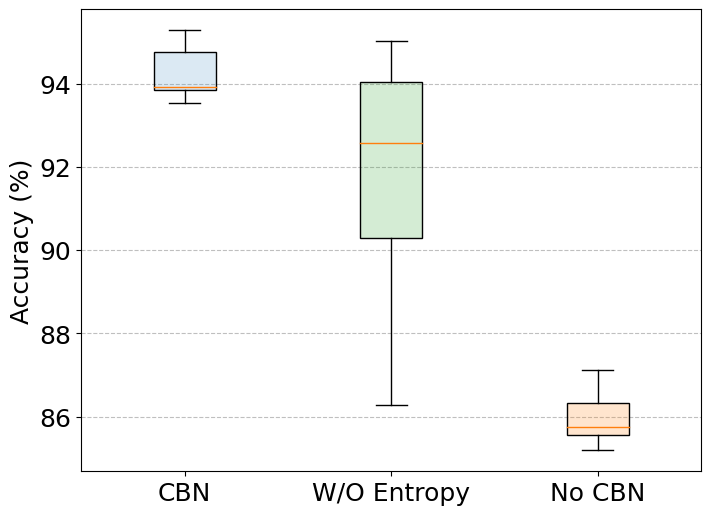}
    \caption{Stability analysis of EverAdapt}
    \label{fig:Stable_study}
     \end{subfigure}

\caption{Model Analysis for Everadapt}
\label{Fig:universal}
\end{figure*}

\subsection{Baseline methods}
To evaluate the performance of our model, we compare it We assessed the efficacy of our EverAdapt technique by comparing it with recent domain adaptation methods proposed for fault diagnosis. We re-implement all the baselines in our framework, while ensuring the same backbone network and training schemes. Overall, the compared methods are as follows: 
\begin{itemize}
    \item Conditional adversarial DA with discrimination embedding (CADA-DE) \cite{yu2020conditional}: utilized a conditional adversarial alignment by integrating task-specific knowledge with the features during the alignment step for the different domains.
    \item Hierarchical deep domain adaptation (HDDA) \cite{wang2019hierarchical}: aligns the second-order statistics of the source and target distributions in order to effectively minimize the shift between the two domains.
    \item Improved Domain Adversarial Neural Network (IDANN) \cite{zhang2022multi}: leverages gradient reversal layer to adversarially train a domain discriminator network against an encoder network.
    \item Minimum Discrepancy Estimation for Deep Domain Adaptation (MMDA) \cite{azamfar2020deep}: combines the MMD and correlation alignment with entropy minimization to effectively address the domain shift issue.
    \item Subdomain adaptation transfer learning network (SATLN) \cite{wang2021subdomain}: leverages gradient reversal layer to adversarially train a domain discriminator network against an encoder network.
\end{itemize}

In addition to the leading domain adaptation methods, we've assessed EverAdapt against CDA methods proposed in other fields which includes:

\begin{itemize}
    \item Continuous unsupervised adaptation (CUA) \cite{bobu2018adapting}: leverage replay sample loss to address catastrophic forgetting
    \item Continual Unsupervised Domain Adaptation (CONDA) \cite{taufique2021conda} build upon the work of \cite{liang2020we} by incorporating it with sample replay with an appropriate sample replay manager to append new target domain samples with class-representative samples.
    \item (DCTLN-DWA) \cite{li2023deep}: combines techniques from adversarial domain adaptation and replay sample loss, which are selected through a herding algorithm to obtain class-representative samples. 
\end{itemize}

\subsection{Comparison with baselines}
We evaluated EverAdapt's performance against various established domain adaptation methods, utilizing both the PU datasets and UO dataset. The comparative results, presented in Table \ref{tab:PU baselines}, are averaged over three distinct scenarios. We found that EverAdapt demonstrated state-of-the-art performance for all datasets, achieving the highest accuracy and the BWT scores across all three datasets while merely trailing behind in adaptation performance. Specifically,
\begin{itemize}
    \item PU Artificial Dataset: EverAdapt demonstrated superior accuracy, outperforming the best baseline methods by 8.67\%. It also led in BWT scores by 1.29\%. In terms of adaptation performance, EverAdapt was ahead by 2.17\%.
    \item PU Real Dataset: EverAdapt exceeded the top baseline methods in accuracy by 2.36\% and in BWT scores by 0.48\%. However, it lagged slightly in adaptation performance, trailing by 0.46\%.
    \item UO Dataset: EverAdapt continued to show excellent performance, surpassing the best baseline methods in accuracy by 0.93\% and in BWT scores by 0.33\%. In adaptation performance, it was behind by 0.94\%.
\end{itemize}

These results indicate that EverAdapt is highly effective in retaining previously learned knowledge while adapting to new tasks which contrasts the baseline CDA methods such as CUA and DCTLN-DWA which demonstrated remarkable BWT scores at the expense of Adapt performance.

%Specifically, EverAdapt outperformed the best baseline methods in terms of accuracy by 8.67\%, 2.36\%, and 0.93\% for the PU Artificial, PU Real, and UO datasets, respectively. This superior performance is attributed to EverAdapt's ability to retain previously learned knowledge without compromising its ability to adapt to new tasks. In terms of BWT, EverAdapt led by 1.29\%, 0.48\%, and 0.33\% for the three datasets, respectively. However, in terms of Adapt, it was ahead by 2.17\% for the PU Artificial dataset but lagged by only 0.46\% and 0.94\% behind the best-performing method for the PU Real and UO datasets, respectively. Contrasting this to CDA methods such as CUA and DCTLN-DWA which demonstrated remarkable BWT scores at the expense of Adapt performance.

This superiority is further illustrated in Figure \ref{fig:First_Target_Forgetting}, which plots the initial target accuracy as the model adapts to various target domains. The plot reveals that our method not only achieves significantly higher initial accuracy, indicating superior adaptation performance but also excels in knowledge retention, as demonstrated by the minimal performance drop compared to other baseline methods. % Detailed performance comparisons for each scenario are available in Tables \ref{tab:PU Artificial baselines} and \ref{tab:PU Real baselines} in the supplementary materials.

%We compared EverAdapt performance against our baseline domain adaptation methods in PU Artificial and PU Real dataset.
%We found that EverAdapt demonstrated state-of-the-art performance for both dataset.
%Table \ref{tab:PU baselines} shows the performance of our method against other baselines for PU real and PU Artificial datasets. 
%The results demonstrate EverAdapt's superior performance in both adaptation and knowledge retention, as evidenced by its higher BWT and ADAPT scores compared to the baseline methods in both datasets. 
%This superiority is further highlighted in Figure \ref{fig:First_Target_Forgetting}, which plots the first target accuracy as the model adapts to target domains.
%In the plot, our method not only shows significantly superior adaptation performance, indicated by higher initial accuracy, but also excels in retaining knowledge. This is evident from the minimal drop in performance compared to other baseline methods

\subsection{Model Analysis}
We conducted an extensive analysis to better understand how our model achieves its state-of-the-art performance.
\subsubsection{Ablation Study}
An ablation study was conducted across three distinct scenarios to assess the efficacy of each component in the EverAdapt model, with results indicating consistent performance improvements in all scenarios. For each scenario, detailed findings are presented in Table \ref{tab:Ablation study}. Initially, the class-conditional alignment exhibited adaptation capabilities but was inadequate in countering catastrophic forgetting. The addition of replay samples improved knowledge retention, enhancing  overall BWT by 7.29\% but slightly reduced adaptation performance by 1.57\%. The integration of CBN significantly boosted both memory retention, with a 7.73\% improvement in BWT, and adaptation performance, improving by 3.16\%. This advancement not only mitigated the initial dip in adaptation performance but also surpassed the performance of the model with only class-conditional alignment by 1.59\%.

\subsubsection{Replay Samples Efficiency}
This study investigates Continual Batch Normalization (CBN)'s role in addressing catastrophic forgetting, focusing on the use of minimal replay sample sizes. In scenario 3 of the PU Artificial dataset, we assessed the effectiveness of preserving merely 1\% of data from each target domain. As illustrated in Figure \ref{fig:Replay_study}, our findings demonstrate CBN's substantial contribution to reinforcing replay sample utility. With just a 1\% replay sample size, CBN notably enhances Backward Transfer (BWT) by nearly 7\%, markedly reducing forgetting to 0.73\%. Furthermore, augmenting the replay size to 10\% while incorporating CBN yields only a slight BWT increment of 0.1\%. This suggests that small replay samples, in conjunction with CBN, effectively combat the catastrophic forgetting challenge.

% This section examines how CBN impacts the efficacy of replay samples in reducing catastrophic forgetting. In parallel, we will be investigating on retaining only 1\% of each target domain compromises performance. This involves evaluating EverAdapt's performance with a focus on varying the replay sample size and the presence or absence of CBN during training.
% Our experiments utilized scenario 3 of the PU Artificial dataset, known for its susceptibility to catastrophic forgetting.

% Our findings, illustrated in Figure \ref{fig:Replay_study}, indicate that CBN enhances the effectiveness of replay samples. Notably, with a 1\% replay sample size, the model employing CBN demonstrates a 5.93\% higher BWT compared to the model without CBN. Furthermore, our results suggest that a 1\% replay size is adequate for mitigating catastrophic forgetting in the PU Artificial dataset, with minimal performance gains observed from further increasing the replay size. At a 10\% replay size with CBN, the improvement in BWT is marginal, increasing only by 0.49\% compared to the model with a 1\% replay size.

\subsubsection{Stability study}
This section presents a stability study of the Continual Batch Normalization (CBN) module within the EverAdapt framework. Focusing on the PU Artificial dataset's scenario 3, we evaluated the significance of individual CBN components in stabilizing the model. Figure \ref{fig:Stable_study} illustrates the performance comparisons between the full implementation of EverAdapt, a variant employing only source statistics normalization without entropy, and another variant excluding CBN entirely. The results affirm the full CBN model's superior performance, indicating the drawbacks of omitting certain components. Specifically, while normalizing target samples with source statistics improved median accuracy by 6.83\%, it also introduced greater variability, evidenced by a fourfold increase in the range of performance outcomes. Integrating entropy, alongside source statistics normalization, significantly counteracted this variability. This emphasizes the critical roles of both entropy incorporation and source normalization in CBN, enhancing not only the model's performance but also its stability under dynamic environmnets.

\section{Conclusion}
In this study, we introduce EverAdapt, a streamlined approach for continual unsupervised domain adaptation in machine fault diagnosis. Central to EverAdapt is the novel Continual Batch Normalization (CBN) technique, which effectively preserves model performance across varying domains and mitigates catastrophic forgetting. By standardizing batch statistics and reducing reliance on extensive replay samples, CBN emerges as the pivotal contribution of this work, ensuring robust and efficient adaptation in dynamic environments. Empirically, EverAdapt has demonstrated superior performance, setting new benchmarks on two real-world datasets, and fostering more robust and practical solutions in the face of dynamic real-world scenarios.

\bibliographystyle{utils/IEEEtrans}
\bibliography{main}

\begin{thebibliography}{10}
\providecommand{\url}[1]{#1}
\csname url@samestyle\endcsname
\providecommand{\newblock}{\relax}
\providecommand{\bibinfo}[2]{#2}
\providecommand{\BIBentrySTDinterwordspacing}{\spaceskip=0pt\relax}
\providecommand{\BIBentryALTinterwordstretchfactor}{4}
\providecommand{\BIBentryALTinterwordspacing}{\spaceskip=\fontdimen2\font plus
\BIBentryALTinterwordstretchfactor\fontdimen3\font minus \fontdimen4\font\relax}
\providecommand{\BIBforeignlanguage}[2]{{%
\expandafter\ifx\csname l@#1\endcsname\relax
\typeout{** WARNING: IEEEtran.bst: No hyphenation pattern has been}%
\typeout{** loaded for the language `#1'. Using the pattern for}%
\typeout{** the default language instead.}%
\else
\language=\csname l@#1\endcsname
\fi
#2}}
\providecommand{\BIBdecl}{\relax}
\BIBdecl

\bibitem{diff_machines_da}
L.~Guo, Y.~Lei, S.~Xing, T.~Yan, and N.~Li, ``Deep convolutional transfer learning network: A new method for intelligent fault diagnosis of machines with unlabeled data,'' \emph{IEEE Transactions on Industrial Electronics}, vol.~66, no.~9, pp. 7316--7325, 2018.

\bibitem{cheng2020wasserstein}
C.~Cheng, B.~Zhou, G.~Ma, D.~Wu, and Y.~Yuan, ``Wasserstein distance based deep adversarial transfer learning for intelligent fault diagnosis with unlabeled or insufficient labeled data,'' \emph{Neurocomputing}, vol. 409, pp. 35--45, 2020.

\bibitem{fine_tune_fd}
S.~Shao, S.~McAleer, R.~Yan, and P.~Baldi, ``Highly accurate machine fault diagnosis using deep transfer learning,'' \emph{IEEE Transactions on Industrial Informatics}, vol.~15, no.~4, pp. 2446--2455, 2019.

\bibitem{sae_da}
L.~Wen, L.~Gao, and X.~Li, ``A new deep transfer learning based on sparse auto-encoder for fault diagnosis,'' \emph{IEEE Transactions on systems, man, and cybernetics: systems}, vol.~49, no.~1, pp. 136--144, 2017.

\bibitem{ae_da1}
W.~Lu, B.~Liang, Y.~Cheng, D.~Meng, J.~Yang, and T.~Zhang, ``Deep model based domain adaptation for fault diagnosis,'' \emph{IEEE Transactions on Industrial Electronics}, vol.~64, no.~3, pp. 2296--2305, 2017.

\bibitem{c_forget}
\BIBentryALTinterwordspacing
J.~Kirkpatrick, R.~Pascanu, N.~Rabinowitz, J.~Veness, G.~Desjardins, A.~A. Rusu, K.~Milan, J.~Quan, T.~Ramalho, A.~Grabska-Barwinska, D.~Hassabis, C.~Clopath, D.~Kumaran, and R.~Hadsell, ``Overcoming catastrophic forgetting in neural networks,'' \emph{Proceedings of the National Academy of Sciences}, vol. 114, no.~13, pp. 3521--3526, 2017. [Online]. Available: \url{https://www.pnas.org/doi/abs/10.1073/pnas.1611835114}
\BIBentrySTDinterwordspacing

\bibitem{conda}
\BIBentryALTinterwordspacing
A.~M.~N. Taufique, C.~S. Jahan, and A.~E. Savakis, ``Continual unsupervised domain adaptation in data-constrained environments,'' \emph{IEEE Transactions on Artificial Intelligence}, vol.~5, pp. 167--178, 2024. [Online]. Available: \url{https://api.semanticscholar.org/CorpusID:255642695}
\BIBentrySTDinterwordspacing

\bibitem{grcl}
S.~Tang, P.~Su, D.~Chen, and W.~Ouyang, ``Gradient regularized contrastive learning for continual domain adaptation,'' in \emph{Proceedings of the AAAI Conference on Artificial Intelligence}, vol.~35, 2021, pp. 2665--2673.

\bibitem{gfr}
\BIBentryALTinterwordspacing
Q.~Lao, X.~Jiang, M.~Havaei, and Y.~Bengio, ``A two-stream continual learning system with variational domain-agnostic feature replay,'' \emph{IEEE Transactions on Neural Networks and Learning Systems}, vol.~33, pp. 4466--4478, 2021. [Online]. Available: \url{https://api.semanticscholar.org/CorpusID:232113812}
\BIBentrySTDinterwordspacing

\bibitem{lu2016deep}
W.~Lu, B.~Liang, Y.~Cheng, D.~Meng, J.~Yang, and T.~Zhang, ``Deep model based domain adaptation for fault diagnosis,'' \emph{IEEE Transactions on Industrial Electronics}, vol.~64, no.~3, pp. 2296--2305, 2016.

\bibitem{li2019deep}
X.~Li, W.~Zhang, N.-X. Xu, and Q.~Ding, ``Deep learning-based machinery fault diagnostics with domain adaptation across sensors at different places,'' \emph{IEEE Transactions on Industrial Electronics}, vol.~67, no.~8, pp. 6785--6794, 2019.

\bibitem{wang2021subdomain}
Z.~Wang, X.~He, B.~Yang, and N.~Li, ``Subdomain adaptation transfer learning network for fault diagnosis of roller bearings,'' \emph{IEEE Transactions on Industrial Electronics}, vol.~69, no.~8, pp. 8430--8439, 2021.

\bibitem{zhu2020new}
J.~Zhu, N.~Chen, and C.~Shen, ``A new multiple source domain adaptation fault diagnosis method between different rotating machines,'' \emph{IEEE Transactions on Industrial Informatics}, vol.~17, no.~7, pp. 4788--4797, 2020.

\bibitem{cua}
\BIBentryALTinterwordspacing
A.~Bobu, E.~Tzeng, J.~Hoffman, and T.~Darrell, ``Adapting to continuously shifting domains,'' 2018. [Online]. Available: \url{https://openreview.net/forum?id=BJsBjPJvf}
\BIBentrySTDinterwordspacing

\bibitem{frida}
\BIBentryALTinterwordspacing
S.~Rakshit, A.~Mohanty, R.~Chavhan, B.~Banerjee, G.~Roig, and S.~Chaudhuri, ``Frida - generative feature replay for incremental domain adaptation,'' \emph{Comput. Vis. Image Underst.}, vol. 217, p. 103367, 2021. [Online]. Available: \url{https://api.semanticscholar.org/CorpusID:245537652}
\BIBentrySTDinterwordspacing

\bibitem{gsfda}
\BIBentryALTinterwordspacing
S.~Yang, Y.~Wang, J.~van~de Weijer, L.~Herranz, and S.~Jui, ``Generalized source-free domain adaptation,'' \emph{2021 IEEE/CVF International Conference on Computer Vision (ICCV)}, pp. 8958--8967, 2021. [Online]. Available: \url{https://api.semanticscholar.org/CorpusID:236881316}
\BIBentrySTDinterwordspacing

\bibitem{packnet}
\BIBentryALTinterwordspacing
A.~Mallya and S.~Lazebnik, ``Packnet: Adding multiple tasks to a single network by iterative pruning,'' \emph{2018 IEEE/CVF Conference on Computer Vision and Pattern Recognition}, pp. 7765--7773, 2017. [Online]. Available: \url{https://api.semanticscholar.org/CorpusID:35249701}
\BIBentrySTDinterwordspacing

\bibitem{eata}
\BIBentryALTinterwordspacing
S.~Niu, J.~Wu, Y.~Zhang, Y.~Chen, S.~D. Zheng, P.~Zhao, and M.~Tan, ``Efficient test-time model adaptation without forgetting,'' in \emph{International Conference on Machine Learning}, 2022. [Online]. Available: \url{https://api.semanticscholar.org/CorpusID:247996873}
\BIBentrySTDinterwordspacing

\bibitem{yu2020conditional}
X.~Yu, Z.~Zhao, X.~Zhang, C.~Sun, B.~Gong, R.~Yan, and X.~Chen, ``Conditional adversarial domain adaptation with discrimination embedding for locomotive fault diagnosis,'' \emph{IEEE Transactions on Instrumentation and Measurement}, vol.~70, pp. 1--12, 2020.

\bibitem{zhang2022multi}
D.~Zhang and L.~Zhang, ``A multi-feature fusion-based domain adversarial neural network for fault diagnosis of rotating machinery,'' \emph{Measurement}, vol. 200, p. 111576, 2022.

\bibitem{wang2019hierarchical}
X.~Wang, H.~He, and L.~Li, ``A hierarchical deep domain adaptation approach for fault diagnosis of power plant thermal system,'' \emph{IEEE Transactions on Industrial Informatics}, vol.~15, no.~9, pp. 5139--5148, 2019.

\bibitem{azamfar2020deep}
M.~Azamfar, X.~Li, and J.~Lee, ``Deep learning-based domain adaptation method for fault diagnosis in semiconductor manufacturing,'' \emph{IEEE Transactions on Semiconductor Manufacturing}, vol.~33, no.~3, pp. 445--453, 2020.

\bibitem{taufique2021conda}
A.~M.~N. Taufique, C.~S. Jahan, and A.~Savakis, ``Conda: Continual unsupervised domain adaptation,'' \emph{arXiv preprint arXiv:2103.11056}, 2021.

\bibitem{bobu2018adapting}
A.~Bobu, E.~Tzeng, J.~Hoffman, and T.~Darrell, ``Adapting to continuously shifting domains,'' 2018.

\bibitem{li2023deep}
J.~Li, R.~Huang, Z.~Chen, G.~He, K.~C. Gryllias, and W.~Li, ``Deep continual transfer learning with dynamic weight aggregation for fault diagnosis of industrial streaming data under varying working conditions,'' \emph{Advanced Engineering Informatics}, vol.~55, p. 101883, 2023.

\bibitem{zhao2020deep}
Z.~Zhao, T.~Li, J.~Wu, C.~Sun, S.~Wang, R.~Yan, and X.~Chen, ``Deep learning algorithms for rotating machinery intelligent diagnosis: An open source benchmark study,'' \emph{ISA transactions}, vol. 107, pp. 224--255, 2020.

\bibitem{lessmeier2016condition}
C.~Lessmeier, J.~K. Kimotho, D.~Zimmer, and W.~Sextro, ``Condition monitoring of bearing damage in electromechanical drive systems by using motor current signals of electric motors: A benchmark data set for data-driven classification,'' in \emph{PHM Society European Conference}, vol.~3, 2016.

\bibitem{uo_dataset}
H.~Huang and N.~Baddour, ``Bearing vibration data collected under time-varying rotational speed conditions,'' \emph{Data in brief}, vol.~21, pp. 1745--1749, 2018.

\bibitem{ragab2023adatime}
M.~Ragab, E.~Eldele, W.~L. Tan, C.-S. Foo, Z.~Chen, M.~Wu, C.-K. Kwoh, and X.~Li, ``Adatime: A benchmarking suite for domain adaptation on time series data,'' \emph{ACM Transactions on Knowledge Discovery from Data}, vol.~17, no.~8, pp. 1--18, 2023.

\bibitem{liang2020we}
J.~Liang, D.~Hu, and J.~Feng, ``Do we really need to access the source data? source hypothesis transfer for unsupervised domain adaptation,'' in \emph{International conference on machine learning}.\hskip 1em plus 0.5em minus 0.4em\relax PMLR, 2020, pp. 6028--6039.

\end{thebibliography}

\end{document}

\maketitle

\section{Paderborn University (PU) Dataset}
The Paderborn University Bearing Data Centre provides a dataset that is an extensive compilation of vibration and current signals recorded from an electric motor, as detailed in \cite{lessmeier2016condition}. This dataset encompasses 32 unique sets of signals, corresponding to 32 distinct bearings, each labeled according to its condition: 6 are healthy, 12 have been artificially damaged, and 14 exhibit real damages incurred under actual working conditions. Measurements were taken for each bearing across four different operational conditions, which are used to create distinct domains within the dataset. Thus, each bearing's data not only serves as a class label indicating its condition but also corresponds to one of the operational domains, providing a multifaceted view of the bearings' performance. Data segmentation, as suggested in the literature \cite{zhao2020deep}, effectively enlarges the dataset and simplifies the model's input requirements. We employed a moving window technique with a window size and stride length of 1024 to segment the data, guaranteeing distinct, non-overlapping data segments for model training. 

\subsection{PU Artificial Dataset}
Table \ref{tab:PU_artificial} presents the details of the artificially damaged bearings dataset. This dataset is divided into eight distinct classes, each representing a unique bearing condition characterized by a combination of damage level, type, and location. Class 1 represents an undamaged bearing, serving as a control for comparison. Classes 2 through 8 describe bearings with different artificial damages like EDM, engraving, and drilling. These damages are categorized by their location on the bearing and their severity, on a scale from 1 to 2. Each class is defined by a unique damage code that reflects these damage parameters

\subsection{PU Real Dataset}
Table \ref{tab:PU_real} showcases the real damages dataset.  Within this dataset, there are five classes, each corresponding to a distinct bearing condition encountered in operational settings. Each class is characterized not just by the damage type but also by the specific location of the damage, either on the inner or outer race. These distinctions are critical as they provide insights into the bearing's operational vulnerabilities and the consequent failure mechanisms.

\subsection{Working Conditions}
The PU datasets, which include data for bearings with both artificial and real damages, features four distinct working conditions to simulate different operational scenarios. These conditions, detailed in Table \ref{tab:PU_working_condition}, are identified as A1/R1 through A4/R4, where 'A' represents the artificial dataset and 'R' the real dataset conditions. The working conditions are defined by variations in rotating speed, load torque, and radial force. For example, A1/R1 and A3/R3 maintain a high rotating speed of 1500 rpm, but differ in load torque and radial force, while A2/R2 operates at a lower speed of 900 rpm.
\section{Baseline Comparison}

Tables \ref{tab:PU Artificial baselines} and \ref{tab:PU Real baselines} present the performance of the proposed method compared to other baselines on the PU Artificial and PU Real datasets, respectively. The evaluation spans three distinct scenarios, considering metrics such as ACC, BWT, and Adapt. The results demonstrate that our method not only consistently surpasses all baselines across these metrics but also exhibits robustness against domain order variations in different scenarios
\begin{table}
  \centering
  \caption{Different Classes for PU Artificial Dataset}
  \label{tab:PU_artificial}
    \resizebox{\columnwidth}{!}{
  \begin{tabular}{lcccccc}
    \toprule
    Classes & Bearing & Damage level & Damage type & Location & Damage code \\
    \midrule
    1 & K001 & 0 & None & N/A & No Damage \\
    2 & KA01 & 1 & EDM & Outer & O-L1-EDM \\
    3 & KA03 & 2 & Engraving & Outer & O-L2-Engraving \\
    4 & KA05 & 1 & Engraving & Outer & O-L1-Engraving \\
    5 & KA07 & 1 & Drilling & Outer & O-L1-Drilling \\
    6 & KI01 & 1 & EDM & Outer & O-L1-EDM \\
    7 & KI03 & 1 & Engraving & Inner & I-L1-Engraving \\
    8 & KI07 & 2 & Engraving & Inner & I-L2-Engraving \\
    \bottomrule
  \end{tabular}}
\end{table}

\begin{table}
  \centering
  \caption{Different Classes for PU Real Dataset}
  \label{tab:PU_real}
  \resizebox{0.5\textwidth}{!}{
  \begin{tabular}{lccccc}
    \toprule
    Classes & Bearing & Damage level & Damage type & Location & Damage code \\
    \midrule
    1 & K001 & 0 & None & N/A & No Damage \\
    2 & KA04 & 1 & EDM & Outer & O-L1-EDM \\
    3 & KB23 & 2 & Engraving & Inner & I-L2-Engraving \\
    4 & KB27 & 1 & Engraving & Outer & O-L1-Engraving \\
    5 & KI04 & 1 & Drilling & Inner & I-L1-Drilling \\
    \bottomrule
  \end{tabular}}
\end{table}

\begin{table}
  \centering
  \caption{Four working conditions of PU datasets, A/R denotes domains from PU Artificial and PU Real}
  \label{tab:PU_working_condition}
  \resizebox{0.5\textwidth}{!}{
  \begin{tabular}{ccccc}
    \toprule
    Domain & Rotating speed (rpm) & Load torque (Nm) & Radial force (N) \\
    \midrule
    A1/R1 & 1500 & 0.7 & 1000  \\
    A2/R2 & 900  & 0.7 & 1000\\
    A3/R3 & 1500 & 0.1 & 1000  \\
    A4/R4 & 1500 & 0.7 & 400   \\
    \bottomrule
  \end{tabular}}
\end{table}

\begin{table*}[htbp]
\centering
\caption{Performance of baseline methods and our method on PU Artificial dataset}
\label{tab:PU Artificial baselines}
\resizebox{\textwidth}{!}{
\begin{tabular}{lccccccccc}
\toprule
\textbf{Methods} & \multicolumn{3}{c}{\textbf{Scenario 1}} & \multicolumn{3}{c}{\textbf{Scenario 2}} & \multicolumn{3}{c}{\textbf{Scenario 3}} \\
\cmidrule(r){2-4} \cmidrule(lr){5-7} \cmidrule(lr){8-10} \\
PU Artificial & ACC & BWT & ADAPT & ACC & BWT & ADAPT & ACC & BWT & ADAPT\\
\midrule
CDAN & 81.39 $\pm$ 0.62 & -4.89 $\pm$ 0.61 & 84.65 $\pm$ 0.40 & 81.39 $\pm$ 0.55 & -5.01 $\pm$ 0.66 & -3.37 $\pm$ 2.70 & 80.03 $\pm$ 0.45 & -6.93 $\pm$ 0.73 & 0.02 $\pm$ 0.27\\
DANN & 79.95 $\pm$ 1.74 & -3.28 $\pm$ 4.88 & 82.13 $\pm$ 4.49 & 77.70 $\pm$ 1.82 & -9.41 $\pm$ 5.23 & -18.14 $\pm$ 13.78 & 76.28 $\pm$ 0.58 & -8.33 $\pm$ 4.62 & -3.88 $\pm$ 3.08\\
DeepCORAL & 83.24 $\pm$ 0.06 & -5.49 $\pm$ 2.78 & 86.89 $\pm$ 1.82 & 82.77 $\pm$ 0.48 & -6.66 $\pm$ 2.47 & -1.83 $\pm$ 3.01 & 80.52 $\pm$ 0.81 & -9.61 $\pm$ 2.12 & -0.32 $\pm$ 1.28\\
DSAN & 81.32 $\pm$ 0.31 & -15.88 $\pm$ 1.27 & 91.90 $\pm$ 0.95 & 81.50 $\pm$ 0.26 & -13.32 $\pm$ 1.76 & 90.50 $\pm$ 1.24 & 81.49 $\pm$ 0.24 & -15.56 $\pm$ 1.34 & 91.84 $\pm$ 0.89\\
MMDA & 83.48 $\pm$ 0.30 & -3.23 $\pm$ 3.17 & 85.63 $\pm$ 1.96 & 81.71 $\pm$ 0.41 & -1.71 $\pm$ 4.50 & -9.78 $\pm$ 3.51 & 80.93$\pm$ 0.94 & -3.78 $\pm$ 4.37 & -0.63 $\pm$ 0.84\\ 
\midrule
EverAdapt & 93.11 $\pm$ 2.58 & -1.68 $\pm$ 0.71 & 94.23 $\pm$ 2.24 & 91.56 $\pm$ 0.41 & -1.04 $\pm$ 0.36 & 92.88 $\pm$ 0.35 & 94.07 $\pm$ 0.98 & -0.73 $\pm$ 0.51 & 94.67 $\pm$ 1.01\\
\bottomrule
\end{tabular}}
\end{table*}

\begin{table*}[htbp]
\centering
\caption{Performance of baseline methods and our method on PU Real dataset}
\label{tab:PU Real baselines}
\resizebox{\textwidth}{!}{
\begin{tabular}{lccccccccc}
\toprule
\textbf{Methods} & \multicolumn{3}{c}{\textbf{Scenario 1}} & \multicolumn{3}{c}{\textbf{Scenario 2}} & \multicolumn{3}{c}{\textbf{Scenario 3}} \\
\cmidrule(r){2-4} \cmidrule(lr){5-7} \cmidrule(lr){8-10} \\
PU Real & ACC & BWT & ADAPT & ACC & BWT & ADAPT & ACC & BWT & ADAPT\\
\midrule
CDAN & 87.38 $\pm$ 1.82 & -8.83 $\pm$ 2.22 & 1.56 $\pm$ 3.99 & 88.77 $\pm$ 1.36 & -9.59 $\pm$ 1.36 & 6.77 $\pm$ 4.54 & 86.74 $\pm$ 2.80 & -12.04 $\pm$ 4.36 & -1.46 $\pm$ 1.61 \\
DANN & 90.91 $\pm$ 1.22 & -8.09 $\pm$ 1.83 & 11.17 $\pm$ 3.00 & 92.24 $\pm$ 1.21 & -4.99 $\pm$ 2.85 & -3.38 $\pm$ 8.71 & 91.27 $\pm$ 0.72 & -7.50 $\pm$ 2.46 & -0.85 $\pm$ 0.46 \\
DeepCORAL & 90.30 $\pm$ 2.00 & -9.13 $\pm$ 3.35 & 3.44 $\pm$ 8.19 & 89.97 $\pm$ 2.23 & -8.59 $\pm$ 1.80 & 3.16 $\pm$ 8.68 & 86.83 $\pm$ 2.93 & -14.51 $\pm$ 5.10 & 1.65 $\pm$ 4.00\\
DSAN & 94.48 $\pm$ 0.72 & -7.03 $\pm$ 1.18 & 99.17 $\pm$ 0.23 & 94.55 $\pm$ 0.83 & -6.42 $\pm$ 0.99 & 98.83 $\pm$ 0.31 & 93.40 $\pm$ 1.10 & -8.65 $\pm$ 1.48 & 99.17 $\pm$ 0.22\\
MMDA & 95.05 $\pm$ 0.42 & -3.34 $\pm$ 0.92 & 0.66 $\pm$ 6.75 & 95.20 $\pm$ 0.45 & -0.80 $\pm$ 2.43 & -8.14 $\pm$ 8.16 & 92.66 $\pm$ 1.36 & -7.02 $\pm$ 2.24 & -0.26 $\pm$ 0.34\\ 
\midrule
EverAdapt & 99.16 $\pm$ 0.30 & 0.21 $\pm$ 0.47 & 99.08 $\pm$ 0.34 & 98.71 $\pm$ 0.89 & -0.13 $\pm$ 0.59 & 98.90 $\pm$ 0.42 & 99.26 $\pm$ 0.27 & 0.33 $\pm$ 0.39 & 99.11 $\pm$ 0.34\\
\bottomrule
\end{tabular}}
\end{table*}

\bibliographystyle{utils/named}
\bibliography{utils/ref}